\documentclass{MITcsail}
\usepackage{times}

\usepackage[colorlinks=true, citecolor=black, linkcolor=black]{hyperref}
\usepackage{url}
\usepackage[utf8]{inputenc} 
\usepackage{booktabs}       
\usepackage{amsfonts}       
\usepackage{nicefrac}       
\usepackage{microtype}      
\usepackage{xcolor}         
\usepackage{bm}
\usepackage{amssymb}
\usepackage{amsmath}
\usepackage{graphicx}
\usepackage{epstopdf}
\usepackage{indentfirst}
\usepackage{wrapfig}
\usepackage{subfig}
\usepackage{array}
\usepackage{color}
\usepackage{booktabs}
\usepackage{adjustbox}
\usepackage{multirow}
\usepackage{multicol}

\title{Are All Vision Models Created Equal? A Study of the Open-Loop to Closed-Loop Causality Gap}

\author
{Mathias Lechner~\footnote{Correspondence E-mail: mlechner@mit.edu}$^{1}$, Ramin Hasani~$^{1}$, Alexander Amini~$^{1}$, Tsun-Hsuan Wang~$^{1}$, \\ Thomas Henzinger~$^{2}$, and Daniela Rus~$^{1}$\\
\vspace{1em} 
\normalfont{\small $^{1}$Massachusetts Institute of Technology (MIT)} \\
\normalsize{\small $^{2}$Institute of Science and Technology Austria (IST Austria)}\\
}

\begin{document}

\maketitle
\thispagestyle{firstpagestyle} 

\begin{abstract}
There is an ever-growing zoo of modern neural network models that can efficiently learn end-to-end control from visual observations. 
These advanced deep models, ranging from convolutional to patch-based networks, have been extensively tested on offline image classification and regression tasks.
In this paper, we study these vision architectures with respect to the open-loop to closed-loop causality gap, i.e., offline training followed by an online closed-loop deployment. This causality gap typically emerges in robotics applications such as autonomous driving, where a network is trained to imitate the control commands of a human. In this setting, two situations arise: 1) Closed-loop testing in-distribution, where the test environment shares properties with those of offline training data. 2) Closed-loop testing under distribution shifts and out-of-distribution. 
Contrary to recently reported results, we show that \emph{under proper training guidelines}, all vision models perform indistinguishably well on in-distribution deployment, resolving the causality gap. In situation 2, We observe that the causality gap disrupts performance regardless of the choice of the model architecture. Our results imply that the causality gap can be solved in situation one with our proposed training guideline with \emph{any} modern network architecture, whereas achieving out-of-distribution generalization (situation two) requires further investigations, for instance, on data diversity rather than the model architecture.
\end{abstract}

\section{Introduction}\label{sec:intro}

\begin{wrapfigure}[11]{r}{0.4\textwidth}
\vspace{-4mm}
	\centering
	\href{https://youtu.be/0GxKzv5Ej88}{\includegraphics[scale=0.36]{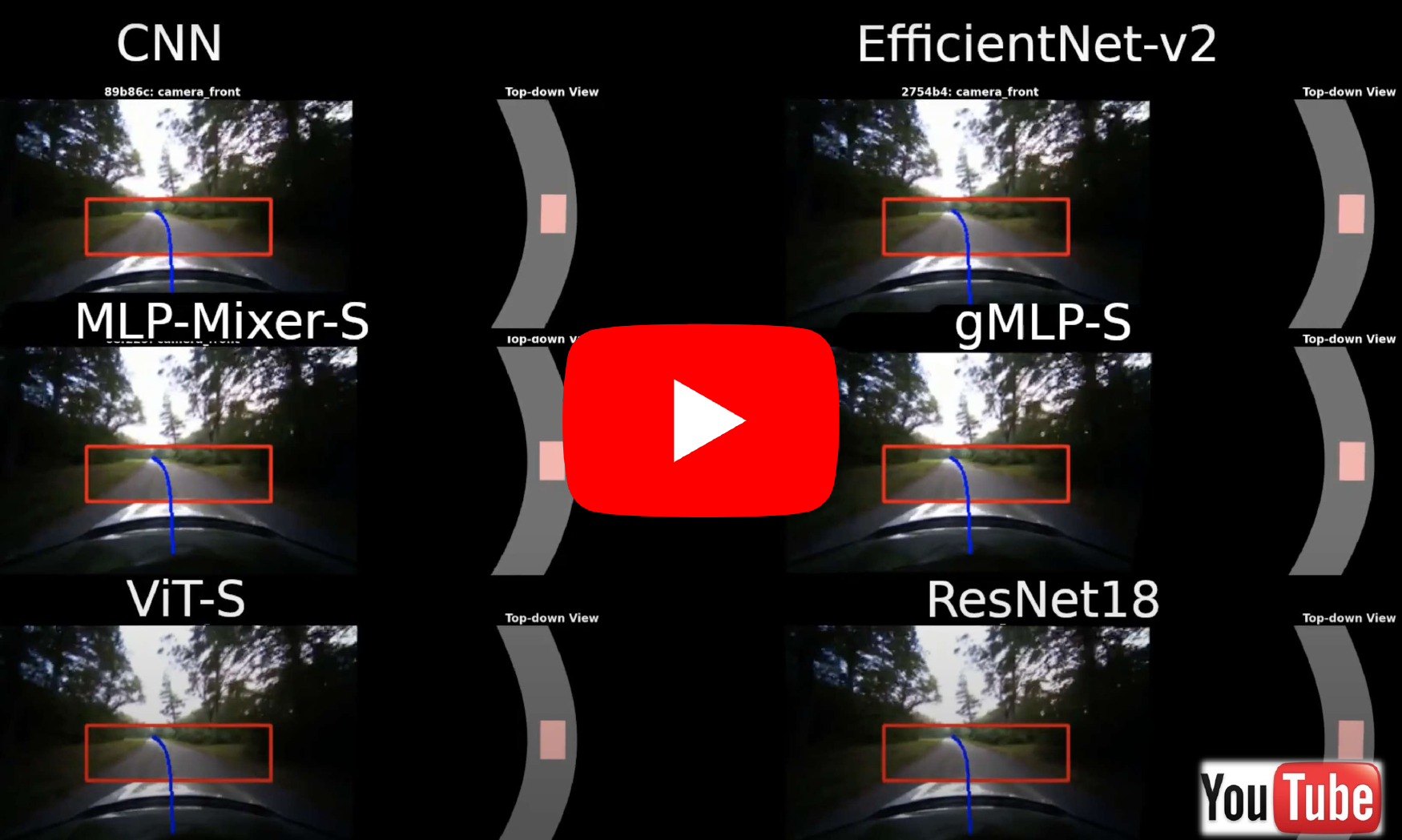}}
	\caption*{Video demonstration of vision models in self driving. \url{https://youtu.be/0GxKzv5Ej88}}%
	\label{fig:teaser}%
\end{wrapfigure}
The deployment of end-to-end learning systems in robotics applications is increasing due to their ability to efficiently and automatically learn representations from high-dimensional observations such as visual inputs without the need for hand-crafting features, bypassing perception to control. 

In this space, a tremendous number of advanced deep learning models have been proposed to perform competitively in end-to-end perception-to-control tasks. For example, patch-based vision architectures such as Vision Transformer (ViT) \citep{dosovitskiy2020image} have shown to be competitive with models based on convolutional neural networks (CNNs) \citep{fukushima1982neocognitron,lecun1989backpropagation,lechner2022entangled} in computer vision applications for which CNNs were the predominant choice. A recent line of research, namely the MLPMixer \citep{tolstikhin2021mlp}, and ConvMixer \citep{trockman2022patches} suggested that the great generalization performance of ViT might be rooted in the patch structure of the inputs rather than the choice of the architecture. There are also works suggesting that self-attention is not crucial in vision Transformers and simply a gating projection in multi-layer perceptrons (MLPs)  \citep{liu2021pay} or replacing self-attention sublayer with an unparameterized Fourier Transform \citep{lee2021fnet} can outperform ViT.

These proposals are largely tested in offline settings where the output decisions of the network do not change the next incoming inputs. In other words, patch-based and mixer models trained offline have not yet been evaluated in a closed-loop with an environment where network actions affect next observations, such as in imitation learning tasks. IL agents typically suffer from a causality gap arising from the transfer of models from open-loop training to closed-loop testing. In this paper, we focus on investigating this gap in a systematic way.

In closed-loop testing, we need to be cognizant of two modes: 1) Closed-loop testing in-distribution. In this setting, we test networks in environments that share similar properties to that of the training environment. 2) Closed-loop testing under distribution shifts and out-of-distribution. 

\begin{figure}[t]
    \centering
    \includegraphics[width=0.45\textwidth]{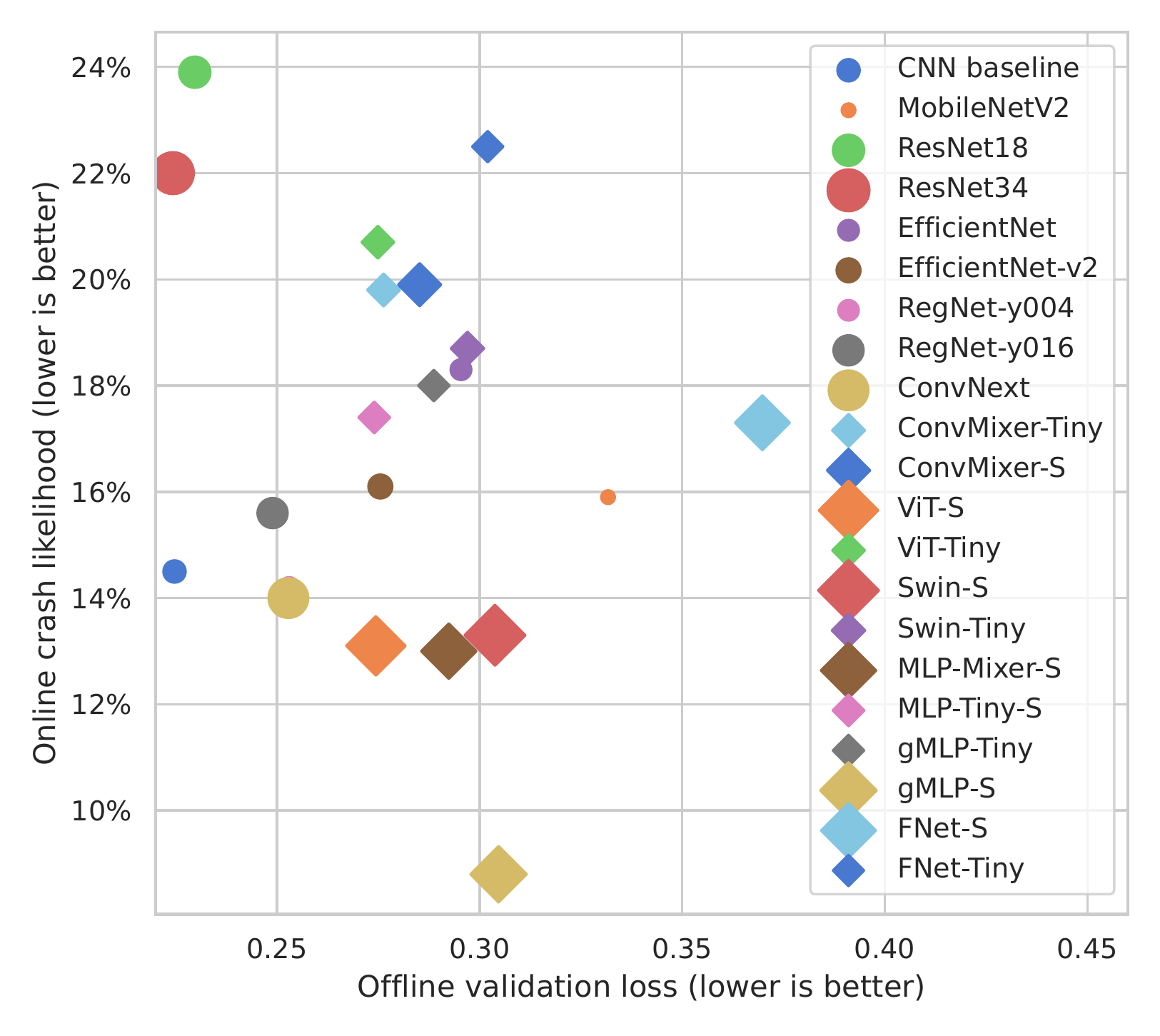}
    \caption{Online deployment vs. offline training causality gap in perspective. Marker size is linearly proportional to the number of trainable parameters.}
    \label{fig:my_label}
\end{figure}

Testing models under both settings requires us to ensure fairness and proper evaluation of the effectiveness of different model architectures in learning robust perception-to-control instances. To this end, we must validate that all baseline models are trained to their best capability given the same amount of hyperparameter optimization budget under a controlled training pipeline.

In an attempt to make a unified evaluation of advanced deep learning models in robot imitation learning tasks, in this paper, we set out to design a series of imitation learning pipelines to train models in a controlled and fair setting and test their generalization capability in and out of distribution.

In particular, we design an end-to-end autonomous driving (AD) IL pipeline based on a photorealistic AD simulation platform called VISTA \citep{amini2020learning}, which can test agents in a closed-loop AD environment synthesizing novel views to assess their closed-loop generalization capabilities \citep{xiao2021barriernet,xiao2022differentiable}. 

Counterintuitively and in contrast to the recently reported results \citep{paul2022vision,naseer2021intriguing,bai2021transformers}, we show that no new architecture is needed to bridge the causality gap between offline training and online testing in-distribution, as our controlled training pipeline enables all models to perform remarkably well on the given tasks. Moreover, for achieving out-of-distribution generalization, we observe that the causality gap certainly affects the performance of models, again, almost regardless of the choice of their architecture. These findings suggest the rethinking of the emphasis on the choice of popular models such as Transformers over CNNs, as other factors such as proper training setup, augmentation strategies, and data diversity play a more important role in generalization in and out of distribution.

To validate our results and make sure our conclusions are not specific to the AD domain, we further extended our experiments (offline training, then online testing) to more standard visual behavior cloning benchmarks such as perception-to-control arcade learning environment (ALE) \citep{bellemare2013arcade} tasks. Similar conclusions are then drawn in this case.

\noindent Our contributions are summarized below:

\begin{enumerate}
\item The design of a unified end-to-end training infrastructure for a fair comparison of advanced deep learning architectures in robot imitation learning applications
    \item Studying patch-based architectures and advanced CNNs as agents in closed-loop with their environments via offline training (imitation) and online testing.
    \item Discovering new insights about where patch-based architectures generalize better than CNNs and vice versa in test-cases in and out of distribution
\end{enumerate}

\begin{figure*}
    \centering
    \includegraphics[width=1.0\textwidth]{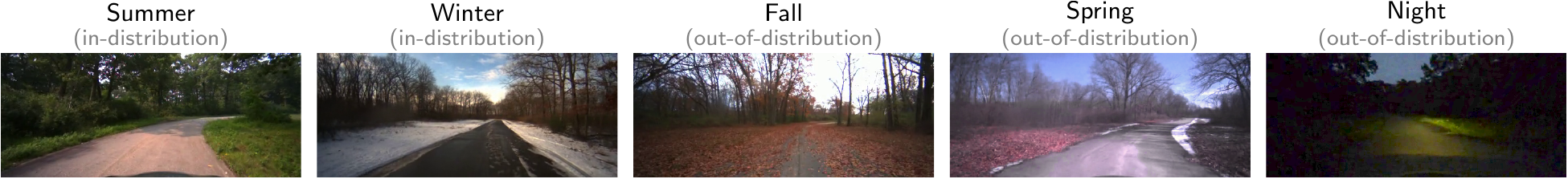}
    \caption{Visualization of sample observations used in our end-to-end AD experiment, spanning across various seasons and times of the day.}
    \label{fig:vistadata}
\end{figure*}

\section{Background and Related Works}
\noindent In this section, we first discuss the image processing architectures studied in this work. Moreover, we recapitulate related works on the understanding of how patch-based CV models process information differently than convolutional architectures. Finally, we discuss existing works on bridging the gap between offline training - online generalization.

\noindent \textbf{Patch-based vision architectures.}
Motivated by the success of Transformers \citep{vaswani2017attention} on natural language processing (NLP) datasets, \citep{dosovitskiy2020image} introduced the \emph{\color{violet}Vision Transformer} (ViT) by adapting the architecture for computer vision tasks. As transformers operate on a 1-dimensional sequence of vectors, \citep{dosovitskiy2020image} proposed to convert an image into a sequence by tiling it into patches. Each patch is then flattened into a vector by concatenating all pixel values.
Researchers have analyzed the difference between how CNNs and ViTs process images \citep{raghu2021vision}. Moreover, it has been claimed that vision transforms are much more robust to image perturbations and occlusions \citep{naseer2021intriguing}, as well as be able to handle distribution-shifts \citep{bai2021transformers,lechner2021adversarial} better than CNNs. However, more recent works have refuted the robustness claims of vision transformers \citep{fu2021patch} by showing that ViTs can be less robust than convolutional networks when considering carefully crafted adversarial attacks.

\noindent \emph{\color{violet}Swin Transformer} \citep{liu2021swin} modifies the vision transformer by adding a hierarchical structure to the feature sequence of patches. The Swin Transformer applies its attention mechanism not to the full sequence but to a window that is shifted over the entire sequence. By increasing network depth, neighboring windows are merged and pooled into large, less fine-grained windows. This hierarchical processing allows it to use smaller patches without exploding the compute and memory footprint of the model.

\noindent \emph{\color{violet}MLP-Mixer} \citep{tolstikhin2021mlp} adapts the idea of vision transformers to map an image to a sequence of patches. This sequence is then processed by alternating plain multi-layer perceptrons (MLP) over the feature and the sequence dimension, i.e., mixing features and mixing spatial information.

\noindent \emph{\color{violet}gMLP} \citep{liu2021pay} is another MLP-only vision architecture that differs from the MLP-Mixer by introducing multiplicative spatial gating units between the alternating spatial and feature MLPs. Empirical results \citep{liu2021pay} show that the gMLP has a better accuracy-parameter ratio than the MLP-Mixer.

\noindent \emph{\color{violet}FNet} \citep{lee2021fnet} replaces the learnable spatial mixing MLP of the MLP-Mixer architecture by a fixed mixing step. In particular, a parameter-free 2-dimensional Fourier transform is applied over the sequence and features dimensions of the input. Although the authors \citep{lee2021fnet} did not evaluate the model for vision tasks, FNet's similarity to patch-based MLP architectures makes it a natural candidate for vision tasks.

\noindent \emph{\color{violet}ConvMixer} \citep{trockman2022patches} replace the MLPs of the MLP-mixer architecture by alternating depth-wise and point-wise 1D convolutions. While an MLP mixes all entries of the spatial and feature dimension, the convolutions of the ConvMixer mix only local information, e.g., kernel size was set to 9 in \citep{trockman2022patches}. The authors claim a large part of the performance of MLP and vision transformers can be attributed to the patch-based processing instead of the type of mixing representation \citep{trockman2022patches}.

\noindent \textbf{Advanced convolutional architectures.}
Here, we briefly discuss modern variants of CNN architectures. 

\noindent \emph{\color{violet}ResNet} \citep{he2016deep} add skip connections that bypass the convolutional layers. This simple modification allows training much deeper networks than a pure sequential composition of layers. Consequently, skips connections can be found in any modern neural network architecture, including patch-based and advanced convolutional models.

\noindent \emph{\color{violet}MobileNetV2} \citep{sandler2018mobilenetv2} replace the standard convolution operations by depth-wise separable convolutions that process the spatial and channel dimension separately. The resulting network requires fewer floating-point operations to compute, which is beneficial for mobile and embedded applications.

\noindent \emph{\color{violet}EfficientNet} \citep{tan2019efficientnet} is an efficient convolutional neural network architecture derived from an automated neural architecture search. The objective of the search is to find a network topology that achieves high performance while simultaneously running efficiently on CPU devices.

\noindent \emph{\color{violet}EfficientNet-v2} fixes the issue of EfficientNets that despite their efficiency on CPU inference, they can be slower than existing architecture types on GPUs at training and inference. 

\noindent \emph{\color{violet}RegNet} \citep{radosavovic2020designing} is a neural network family that systematically explores the design space of previously proposed advances in neural network design. The RegNet-Y subfamily specifically scales the width of the network linearly with depth and comprises squeeze-and-excitation blocks.

\noindent \emph{\color{violet}ConvNext} \citep{liu2022convnet} is a network that subsumes many recent advances in the design of vision architectures, including better activation functions, replacing batch-norm by layer-normalization, and a larger kernel size into standard ResNets.

\noindent \textbf{Imitation learning (IL).} IL describes learning an agent by expert demonstrations consist of observation-action pairs \citep{schaal1999imitation}, directly via behavior cloning \citep{ho2016generative}, or indirectly via inverse reinforcement learning \citep{ng2000algorithms}. When IL agents are deployed online, they most often deviate from the expert demonstrations leading to compounding errors and incorrect inference. Numerous works have tried to address this problem by adding augmentation techniques that collect data from the cloned model in closed-loop settings. This includes methods such as DAgger \citep{ross2010efficient,ross2011reduction}, state-aware imitation \citep{schroecker2017state,le2018hierarchical,desai2020imitation}, pre-trained policies through meta-learning \citep{duan2017one,yu2018one}, min-max optimization schemes \citep{ho2016generative,baram2017end,wu2019imitation,sun2019adversarial}, using insights from causal inference \citep{ortega2021shaking,janner2021reinforcement}, and using a world-model in the loop \citep{ha2018world,amini2022vista,brunnbauer2022latent,ullman2017mind}.

\begin{figure}
    \centering
    \includegraphics[width=0.9\textwidth]{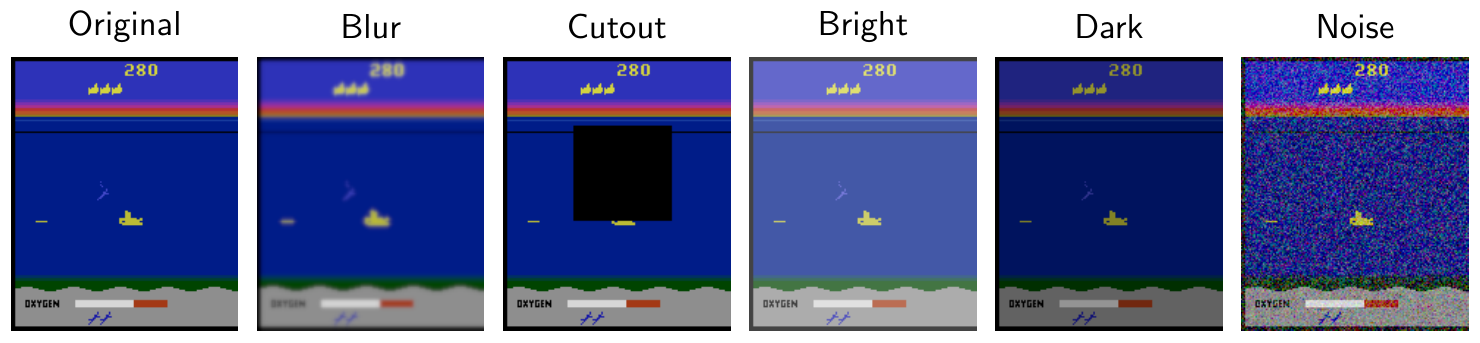}
    \caption{Visualization of the distribution shifts for the Arcade experiments.}
    \label{fig:atariood}
\end{figure}

\noindent \textbf{OOD generalization.} It is fundamentally challenging for statistical models to tackle OOD problems \citep{achille2018information,liang2018enhancing,hendrycks2021natural}, such as domain adaptation \citep{ben2007analysis,muandet2013domain,ganin2016domain,gong2016domain,tzeng2017adversarial}, debiasing \citep{huang2017arbitrary,geirhos2018imagenet,wang2018learning,kim2019learning,clark2019don}, and even practically more challenging settings where OOD semantics are unlabeled \citep{arjovsky2019invariant,rosenfeld2020risks,wang2021causal,krueger2021out}. A large body of recently proposed solutions to OOD generalization, explored causal inference such as causal interventions \citep{wang2021causal,ortega2021shaking}, designing counterfactual schemes \citep{niu2021counterfactual,yue2021counterfactual}, and using attention-based models \citep{kim2018explainable,madan2020fast,goyal2020recurrent,chen2021decision,janner2021reinforcement,yang2021causal}. Here, our study aims to explore how advanced vision networks compare in terms of OOD generalization in online closed-loop with their environments, when trained offline. 

\section{Methodology} 
In this section, we first describe our recipe for how to systematically train end-to-end imitation learning agents offline via a fair hyperparameter tuning pipeline. We then narrate our experimental setup, followed by the method we use for systematic online testing in and out of distribution.

\subsection{Fair Training Setup}
End-to-end deep learning models are typically benchmarked against each other, where one model showed to be outperforming the other. But is it truly the case? Here, we set out to design a controlled offline training to an online testing setup to fairly investigate how advanced vision baselines compare with each other. The training recipe is as follows:

\begin{enumerate}
    \item We conduct a systematic hyperparameter tuning process (described in detail in the next subsection) for each of the 21 tested advanced deep models individually. In particular, We ran a grid search over the two most influential hyperparameters, the learning rate and the weight decay rate. 
    \item We do not perform any early stopping but train a substantial number of optimization steps, which has been shown to be vital for generalization, especially on smaller datasets \citep{power2022grokking,hasani2021closed,vorbach2021causal,lechner2020neural}.
    \item We deploy a custom staircase learning rate decay schedule that decreases the learning rate over the training process by dividing the learning rate by four at 60\%, 80\%, and 93\% of the training epochs.
    \item We warm up the training by running the first epochs with 1/10th of the initial learning rate in order to have the moments' estimates in Adam \citep{kingma2014adam}, Batch-Normalization \citep{ioffe2015batch}, and Layer-Normalization \citep{ba2016layer} modules initialized properly.
    \item We replace the standard Adam optimizer with AdamW \citep{loshchilov2017decoupled}, which decouples the weight decay rate from the loss function, thus avoiding biasing the moments' estimators of Adam.
    \item We apply a rich set of data augmentation techniques, including random brightness, contrast, and saturation modifications, guided policy learning \citep{levine2013guided}, and add noise to the expert's actions during exploration \citep{Hasani2021liquid}.
\end{enumerate}

\noindent\textbf{Baselines} We compare all neural network architectures introduced in the related work section. Overall, we test 21 advanced models, including nine modern convolutional networks and 12 modern patch-based architectures. The exact architecture of these models is directly taken from their reported research article. Additionally, we included a controlled vanilla CNN baseline that comprises seven convolutional layers, each followed by a batch-normalization layer and a ReLU activation function. The first convolution applies a 5-by-5 kernel with 64 filters. The following convolution layers all apply a 3-by-3 kernel with 128, 128, 256, 256, 512, and 512 filters, respectively. A global average pooling layer is applied to feature maps of the final convolution layer, followed by a fully-connected layer with 512 units.

\subsection{Hyperparameter tuning}
In order to have a fair comparison, we perform a systematic hyperparameter tuning process for each architecture. Particularly, we run a grid search over the learning rate and regularization factors (weight decay and dropout rate), which have been shown to have the strongest impact on the performance of the neural networks \citep{krogh1991simple,srivastava2014dropout,loshchilov2017decoupled}.
The objective function of the tuning process is set to the validation loss of the end-to-end driving task. The grid search first searches for the optimal learning rate by evaluating the network with a learning rate of $\{0.01,0.003,0.001,0.0003\}$. Next, the learning rate is fixed to the best performing one, and the search aims to find the right strength of the regularization factor. We evaluate four levels of regularization strengths measured by a pair $(w,d)$, where $w$ is the weight decay factor, and $d$ is the dropout rate applied within the network and before the last layer in each architecture. The grid search evaluates the points $\{(10^{-6},0),(10^{-5},0),(10^{-6},0.2),(10^{-4},0.2)\}$, i.e., spanning from a low regularization pressure to a strong one.

\begin{figure}[t]
    \centering
    \includegraphics[width=0.5\textwidth]{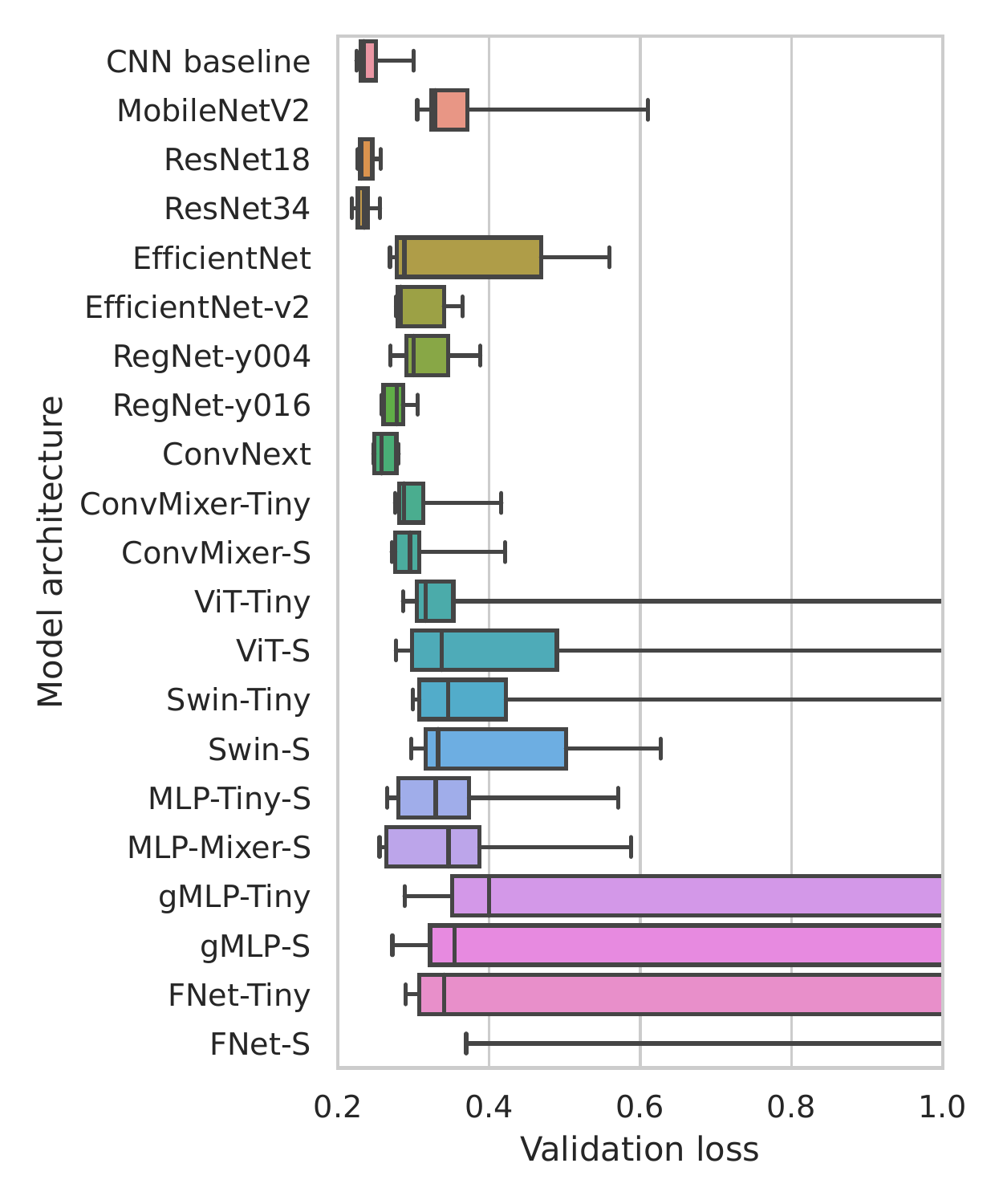}
    \caption{Box-plot showing the validation loss distribution for the different hyperparameters tested for each model. The whiskers represent the minimum/maximum, the box the 0.25, 0.5, and 0.75 quantiles of the values. }
    \label{fig:hp}
\end{figure}

Figure \ref{fig:hp} visualizes the distribution of obtained validation scores of the tested hyperparameters. Most notably, the convolutional architectures tend to have lower variance, i.e., tolerate a wider set of hyperparameters. Moreover, the individual best scores of the models are all in a relatively small range, i.e., between 0.2 and 0.3, demonstrating the necessity of a proper hyperparameter tuning process.

\begin{table*}[t]
    \caption{End-to-end autonomous driving. Numbers show the number of experiment runs that crashed before successful termination. The value in parentheses shows the percentage. The experiments for each model in each column are repeated 200 times. Total number of experiments=21000 (1000 inference experiments for each model in 5 different environments).}
    \centering
        \begin{adjustbox}{width=0.8\textwidth}
    \begin{tabular}{l|cc|lll|c}\toprule
   \textbf{Model} & \multicolumn{6}{c}{\textbf{Number of crashes}} \\\midrule
Condition:  & Summer  & Winter & Fall & Spring & Night & All \\
Seen in training: & \multicolumn{2}{c}{(in-distribution)} & \multicolumn{3}{c}{(out-of-distribution)} &  \\\midrule
CNN baseline & 0 (0\%) & 0 (0\%) & 13 (7\%) & 24 (12\%) & 108 (54\%) & 145 (15\%)\\
MobileNetV2 & 0 (0\%) & 0 (0\%) & 28 (15\%) & 48 (24\%) & 83 (42\%) & 159 (16\%)\\
ResNet18 & 0 (0\%) & 0 (0\%) & 64 (32\%) & 57 (29\%) & 118 (59\%) & 239 (24\%)\\
ResNet34 & 0 (0\%) & 0 (0\%) & 59 (30\%) & 46 (23\%) & 115 (58\%) & 220 (22\%)\\
EfficientNet & 0 (0\%) & 0 (0\%) & 33 (17\%) & 45 (23\%) & 105 (53\%) & 183 (19\%)\\
EfficientNet-v2 & 0 (0\%) & 0 (0\%) & 23 (12\%) & 39 (20\%) & 99 (50\%) & 161 (17\%)\\
RegNet-y004 & 0 (0\%) & 0 (0\%) & 18 (9\%) & 44 (22\%) & 80 (40\%) & 142 (15\%)\\
RegNet-y016 & 0 (0\%) & 0 (0\%) & 12 (6\%) & 48 (24\%) & 96 (48\%) & 156 (16\%)\\
ConvNext & 0 (0\%) & 0 (0\%) & 16 (8\%) & 49 (25\%) & 75 (38\%) & 140 (15\%)\\\midrule
ConvMixer-Tiny & 0 (0\%) & 0 (0\%) & 30 (15\%) & 58 (29\%) & 110 (56\%) & 198 (20\%)\\
ConvMixer-S & 0 (0\%) & 0 (0\%) & 25 (13\%) & 63 (32\%) & 111 (56\%) & 199 (20\%)\\
ViT-S & 0 (0\%) & 0 (0\%) & 21 (11\%) & 40 (20\%) & 70 (35\%) & 131 (14\%)\\
ViT-Tiny & 0 (0\%) & 0 (0\%) & 22 (11\%) & 67 (34\%) & 118 (59\%) & 207 (21\%)\\
Swin-S & 0 (0\%) & 0 (0\%) & 13 (7\%) & 55 (28\%) & 65 (33\%) & 133 (14\%)\\
Swin-Tiny & 0 (0\%) & 0 (0\%) & 23 (12\%) & 65 (33\%) & 99 (50\%) & 187 (19\%)\\
MLP-Mixer-S & 0 (0\%) & 0 (0\%) & 24 (12\%) & 58 (29\%) & 48 (24\%) & 130 (13\%)\\
MLP-Tiny-S & 0 (0\%) & 0 (0\%) & 1 (1\%) & 63 (32\%) & 110 (56\%) & 174 (18\%)\\
gMLP-Tiny & 0 (0\%) & 0 (0\%) & 9 (5\%) & 48 (24\%) & 123 (62\%) & 180 (18\%)\\
gMLP-S & 0 (0\%) & 0 (0\%) & 0 (0\%) & 57 (29\%) & 31 (16\%) & 88 (9\%)\\
FNet-S & 0 (0\%) & 0 (0\%) & 48 (24\%) & 71 (36\%) & 54 (27\%) & 173 (18\%)\\
FNet-Tiny & 0 (0\%) & 0 (0\%) & 29 (15\%) & 63 (32\%) & 133 (67\%) & 225 (23\%)\\
\midrule
Bold threshold & & & $\leq$5\% & $\leq$ 20\% & $\leq$30\% \\
\bottomrule\end{tabular}
\end{adjustbox}
    \label{tab:my_label}
\end{table*}

\begin{table*}[t]
    \centering
    \caption{In-distribution online testing of the Arcade Learning Environments. Numbers report the average episode return for 21 models on 13 different games; each model is trained three times and tested 20 times at inference. Total of 16380 experiments. Values within $1/5$ of the standard deviation of the best performing model are highlighted in bold.}
    \begin{adjustbox}{width=1\textwidth}
    \begin{tabular}{l|ccccccccccccc|c}
      \toprule
 Model  & \rotatebox[origin=c]{65}{Alien} & \rotatebox[origin=c]{65}{Beamrider} & \rotatebox[origin=c]{65}{Breakout} & \rotatebox[origin=c]{65}{Defender} & \rotatebox[origin=c]{65}{Enduro} & \rotatebox[origin=c]{65}{Frogger} & \rotatebox[origin=c]{65}{Frostbite} & \rotatebox[origin=c]{65}{Gopher} & \rotatebox[origin=c]{65}{Pinball} & \rotatebox[origin=c]{65}{Pong} & \rotatebox[origin=c]{65}{Qbert} & \rotatebox[origin=c]{65}{Seaquest} & \rotatebox[origin=c]{65}{Spaceinvaders} & \rotatebox[origin=c]{65}{Ranking}\\\midrule
CNN baseline & 1388 & 6541 & 41 & 10470 & 640 & 544 & \textbf{8026} & 2088 & 14239 & 11 & 12451 & \textbf{4574} & \textbf{824} & 3\\
MobileNetV2 & 1468 & 6282 & 43 & 8448 & 576 & 505 & 7663 & 2335 & 13140 & \textbf{12} & \textbf{12740} & 4124 & 784 & 2\\
ResNet18 & 1454 & \textbf{6818} & \textbf{46} & 10104 & 642 & \textbf{549} & 7486 & 2308 & 18219 & \textbf{13} & \textbf{12986} & \textbf{4483} & 750 & \textbf{6}\\
ResNet34 & \textbf{1547} & 6514 & \textbf{49} & \textbf{11825} & 610 & \textbf{557} & \textbf{8013} & 2187 & 19572 & 12 & 12614 & \textbf{4369} & 766 & \textbf{6}\\
EfficientNet & 1408 & 6406 & \textbf{47} & 9908 & 622 & \textbf{560} & 7490 & 2328 & \textbf{24018} & \textbf{12} & \textbf{12884} & 4106 & 781 & \textbf{5}\\
EfficientNet-v2 & 1475 & 6476 & \textbf{46} & 10168 & 638 & \textbf{563} & \textbf{8251} & 2210 & 19860 & \textbf{12} & \textbf{12785} & 4275 & 756 & \textbf{5}\\
RegNet-y004 & 1499 & \textbf{6630} & 41 & 10848 & 644 & 547 & \textbf{8128} & 2194 & 20970 & 11 & \textbf{12972} & 4301 & \textbf{850} & 4\\
RegNet-y016 & 1444 & 6459 & 43 & 10498 & \textbf{663} & 538 & \textbf{7918} & 2016 & 16847 & 12 & \textbf{12876} & 4185 & \textbf{839} & 4\\
ConvNext & 1459 & 6343 & 45 & \textbf{11420} & 591 & 509 & 6800 & 2143 & 19518 & 12 & \textbf{12696} & 4352 & 769 & 2\\\midrule
ConvMixer-Tiny & \textbf{1515} & 6535 & \textbf{46} & 9396 & 636 & \textbf{556} & 7769 & 2164 & 19842 & 11 & 12610 & 4316 & 798 & 3\\
ConvMixer-S & 1425 & 6363 & 41 & 9190 & 625 & 504 & 6905 & 2327 & 18474 & 11 & 12520 & 4206 & 769 & 0\\
ViT-S & \textbf{1530} & 6137 & 38 & 9256 & 583 & 512 & 7346 & 2168 & 16102 & 11 & \textbf{12649} & \textbf{4516} & \textbf{813} & 4\\
ViT-Tiny & 1442 & \textbf{6684} & 42 & 8705 & 603 & 540 & 7312 & 2031 & 19634 & 12 & 12565 & \textbf{4382} & \textbf{841} & 3\\
Swin-S & \textbf{1538} & 6143 & 33 & 9876 & 518 & 416 & 7297 & 2108 & 21341 & 10 & \textbf{13220} & 4122 & 791 & 2\\
Swin-Tiny & 1473 & \textbf{6630} & 40 & 9174 & 595 & 468 & 7716 & 2128 & 21827 & 12 & 12089 & 4021 & \textbf{843} & 2\\
MLP-Mixer-S & 1465 & 6250 & 38 & 10058 & 607 & 509 & 7251 & 2081 & 18187 & 11 & 12111 & 4301 & 800 & 0\\
MLP-Tiny-S & \textbf{1542} & 5874 & 38 & 10192 & 591 & 513 & 7443 & 2009 & 19518 & 12 & \textbf{12738} & 3992 & 754 & 2\\
gMLP-Tiny & \textbf{1569} & \textbf{6863} & 38 & 10589 & 620 & \textbf{550} & 7634 & 1957 & 17126 & -21 & \textbf{13167} & 4242 & 783 & 4\\
gMLP-S & 1451 & 6520 & 41 & \textbf{11831} & 612 & 544 & 7562 & 2058 & 16642 & -21 & 12584 & 4159 & \textbf{855} & 2\\
FNet-S & 1443 & 6361 & 41 & 9645 & 577 & 486 & 7818 & 2220 & 17883 & \textbf{13} & \textbf{13132} & 3873 & \textbf{887} & 3\\
FNet-Tiny & 1433 & 6290 & 38 & 9570 & 571 & 487 & 7865 & \textbf{2469} & 20219 & 11 & \textbf{12951} & \textbf{4474} & \textbf{832} & 4\\
\midrule Expert policy & 1307 & 6979 & 49 & 10495 & 627 & 482 & 7737 & 2222 & 18318 & 14 & 12995 & 3177 & 699 &  \\
 $\pm$  1/5 std. dev. & $\pm$ 67 & $\pm$ 321 & $\pm$ 3 & $\pm$ 753 & $\pm$ 18 & $\pm$ 15 & $\pm$ 358 & $\pm$ 126 & $\pm$ 1772 & $\pm$ 1 & $\pm$ 583 & $\pm$ 206 & $\pm$ 82 &  \\
\bottomrule
    \end{tabular}
    \end{adjustbox}
    \label{tab:atari}
\end{table*}

\section{Experimental Results}
In this section, we describe our findings on a series of end-to-end imitation learning from offline training to online testing tasks, ranging from autonomous driving to many arcade learning environment games. 
\subsection{End-to-end autonomous driving}
Our first experiment concerns learning the end-to-end control of an autonomous vehicle. We collect data on a full-scale autonomous vehicle with a 30Hz BFS-PGE-23S3C-CS RGB Camera with resolution 960 $\times$ 600 and 130$^\circ$. Each image is temporally synchronized with the steering angle estimated by a differential GPS and an IMU to construct a training pair. The dataset consists of roughly 5-hour driving data collected in different times of the day, different road types, and different seasons, e.g., see Figure \ref{fig:vistadata}. Among all variations, we use summer and winter data for training set with a fraction put aside for (in-distribution) testing and leave fall, spring, and night data for (out-of-distribution) evaluation. For image preprocessing, we perform center cropping as we focus on lane tracking in this work, and we adopt data augmentation, including randomization in brightness, saturation, hue, and gamma, finally followed by per-image normalization. 
To improve over compounding error generated by imitation learning, we use Guided Policy Learning (GPL) \citep{levine2013guided} to generate off-orientation training data and teach the policy how to recover from such scenarios \citep{amini2022vista}. 
To test our model in a closed-loop setting, we leverage a high-fidelity data-driven simulator \citep{amini2022vista} that can be built upon the collected dataset. Trained agents are placed within these simulated environments and are capable of perceiving novel viewpoints in the scene as they execute their policies. The resolution of the input images is 48-by-160 pixels, and all models are trained for 600k steps with a batch size of 64. 

For each model and data condition pair (summer, winter, fall, spring, and night), we run a total of 200 evaluations. An evaluation consists of the model controlling the vehicle's steering with a constant velocity until the vehicle either crashes (i.e., leaves the road) or a certain distance has been driven. We report the number of evaluations that terminated with a crash as our performance metric, with an optimal model counting zero crashes.

The result in Table \ref{tab:my_label} shows the number of crashes for the five different environmental conditions and the aggregated counts over all 1000 evaluation runs. The first two columns show that no crash was observed for any model in the summer and winter conditions. Note that data used for the summer and winter simulation does not overlap with the training data, they only share the season of their data collection process. 
In the out-of-distribution environment conditions, no model was able to maneuver the vehicle across all 600 runs successfully. The best performing model, the gMLP-S had no crash when simulated in fall, but a significant crash rate of 29\% and 16\% in the spring and night conditions, respectively. Figure \ref{fig:my_label} contrasts the offline performance measured by the validation loss on the x-axis with the online performance measured by crash likelihood on the y-axis.
When comparing the convolutional neural network with the patch-based architectures, no significant discrepancy is observed.

\subsection{Arcade learning environment}
As the performance of neural networks is very tasks-specific, a large set of different evaluations are necessary to draw significant conclusions.
Here, we study the offline-online generalization gap of vision architectures on a total of 13 different IL tasks. In particular, we use the Arcade Learning Environment (ALE) \citep{bellemare2013arcade}, which provides a large number of closed-loop environments and images as observations.

The training data for the imitation learning setup are generated by an expert policy in the form of a deep Q-network (DQN) \citep{mnih2015human} from the stable-baselines3 repository \citep{stable-baselines3}. Specifically, we collect the observations and the corresponding suggested action of DQN as labels for the behavior cloning. As regularization, we inject noise into the closed-loop interaction of the data collection process by overwriting 10\% of the DQN's actions with randomly sampled ones. Note that the random actions are only used for driving exploration; the training data contain no random actions.

We collect a total of 2 million data samples for each task, which we split into a training and a validation set with a ratio of 85\%:15\% of non-overlapping trajectories.
We preprocess the observations by converting the images by rescaling them to gray-scale with a resolution of 84-by-84 and stacking the four most recent frames, i.e., a technique known as \emph{deepmind} processor in the literature. For each architecture, we use the tuned hyperparameters from before and train for 200k steps.

\begin{table}[t]
    \centering
        \caption{Mean episode return of the OOD deployment of the Arcade models. Values within $1/5$ of the standard deviation of the best performing model are highlighted in bold.
    }
\begin{tabular}{l|ccccc}\toprule
\toprule
 Model  & Alien & Beamrider & Qbert & Seaquest & Rank\\\midrule
CNN baseline & 877 & 4889 & 8137 & 2906& 0 \\
MobileNetV2 & 922 & 4723 & 6360 & 2218& 0 \\
ResNet18 & 1024 & \textbf{5486} & 7006 & 3219& 1 \\
ResNet34 & 990 & 5167 & 7140 & 3465& 0 \\
EfficientNet & 770 & 4648 & 6665 & 2922& 0 \\
EfficientNet-v2 & 921 & 4917 & 7182 & 3399& 0 \\
RegNet-y004 & 891 & \textbf{5329} & 7001 & 3251& 1 \\
RegNet-y016 & 787 & 4997 & 5847 & 2886& 0 \\
ConvNext & \textbf{1292} & \textbf{5443} & 6970 & 3505& 2 \\\midrule
ConvMixer-Tiny & 1098 & 5101 & 7778 & 1938& 0 \\
ConvMixer-S & 1153 & \textbf{5549} & \textbf{8412} & 2428& 2 \\
ViT-S & 1025 & 4353 & 5983 & 3177& 0 \\
ViT-Tiny & 1044 & 4618 & 5654 & 2280& 0 \\
Swin-S & 1211 & \textbf{5268} & 8023 & 2812& 1 \\
Swin-Tiny & 1188 & 5203 & 6512 & 2670& 0 \\
MLP-Mixer-S & 1082 & 4925 & 6384 & 3030& 0 \\
MLP-Tiny-S & 1141 & 4249 & 5822 & 3060& 0 \\
gMLP-Tiny & 1143 & \textbf{5580} & 7190 & 2742& 1 \\
gMLP-S & 1192 & 5114 & 7017 & \textbf{3586}& 1 \\
FNet-S & 1080 & \textbf{5330} & \textbf{9180} & 3166& 2 \\
FNet-Tiny & 1043 & \textbf{5318} & \textbf{8479} & 3262& 2 \\
\bottomrule
\end{tabular}
    \label{tab:atariood}
\end{table}

Our performance measure is the episodic return, i.e., the non-discounted sum of rewards, when deploying the networks in the closed-loop for a total of 20 episodes. We repeat this experiment, including the training process, for three different random seeds. The results are shown in Table \ref{tab:atari}.
For each environment, we highlight scores that are within 1/5 of the standard deviation of the best-performing model in bold. We also rank the different network architectures by counting for how many environments they are bolding threshold.
The results show that no single model consistently outperforms other architectures and that overall, each model could achieve an excellent closed-loop performance (with only a few outliers). Nonetheless, we observe a minor trend that conv networks achieve a top (bolded) performance more frequently compared to patch-based networks.

\subsection*{Arcade learning environment under distribution-shift}
We also perform an evaluation of how well the models trained on the ALE tasks tolerate a change in the data distribution. 
In particular, we study the effects of 10 different data modification policies on the closed-loop performance measured in the episodic return of the trained ALE models.
The data modification policies are Gaussian blurring with kernel size 3x3 and 5x5, cutting out a 6x6 and 12x12 window in the center of the frame, increasing the brightness by 16 and 32, decreasing the brightness by 16 and 32, and adding uniform noise with a spread of $\pm$ 16 and $\pm$ 32 (pixel values range from 0 to 255). 
We select the ALE environments \textit{Alien},  \textit{Beamrider},  \textit{QBert}, and  \textit{Seaquest} for our evaluation as most models achieve a decent performance on them, thus avoiding skewing the distribution-shift robustness by a difference in in-distribution performance.
Figure \ref{fig:atariood} visualizes the different types of applied data modification policies. 

The results of the ALE out-of-distribution experiments are shown in Table \ref{tab:atariood}. Contrarily to the in-distribution evaluations, we observe a minor trend of the patch-based models performing slightly better than convolutional neural network architectures in environments with shifted distributions.

\section{Conclusion}
In this work, we studied the open-loop to closed-loop causality gap where a neural network is trained offline on labeled data but deployed in a closed-loop system where the decisions of the network affect the next observation, i.e., a setting often found in control and robotics.
Our study focused on image data and modern vision models, specifically, we compared convolutional neural networks with recently proposed patch-based architectures.
Our results showed that if properly trained, any architecture can handle the open-loop to closed-loop causality gap. We also showed a change in the data distribution could have catastrophic consequences on the causality gap. 

Our results indicate that more fundamental approaches than advances in vision processing architectures are necessary to handle the causality gap and a possible change in the data distribution simultaneously. A potential future direction could be to look into incorporating discrete \citep{hasani2016efficient,hasani2017compositional,hasani2019response,goyal2020recurrent} and continuous recurrent mechanisms \citep{lechner2019designing,lechner2020learning,hasani2020natural}, and structural state-space models \citep{lechner2020gershgorin,gu2021efficiently,hasani2022liquid} in the design of our future vision networks.

\section*{ACKNOWLEDGMENT}

This work was partially supported in parts by the ERC-2020-AdG 101020093. Additionally, it was partially sponsored by the United States Air Force Research Laboratory and the United States Air Force Artificial Intelligence Accelerator and was accomplished under Cooperative Agreement Number FA8750-19-2-1000. The views and conclusions contained in this document are those of the authors and should not be interpreted as representing the official policies, either expressed or implied, of the United States Air Force or the U.S. Government. The U.S. Government is authorized to reproduce and distribute reprints for Government purposes notwithstanding any copyright notation herein. This work was further supported by The Boeing Company and the Office of Naval Research (ONR) Grant N00014-18-1-2830.


\end{document}